\PassOptionsToPackage{unicode}{hyperref}
\PassOptionsToPackage{hyphens}{url}

\documentclass[
  twocolumn]{article}
\usepackage{ijcai26}

\usepackage{times}
\usepackage{soul}
\usepackage{url}
\usepackage[hidelinks]{hyperref}
\usepackage[utf8]{inputenc}
\usepackage[small]{caption}
\usepackage{graphicx}
\usepackage{amsmath}
\usepackage{amsthm}
\usepackage{booktabs}
\usepackage{algorithm}
\usepackage{algorithmic}
\usepackage[switch]{lineno}

\usepackage{pifont}
\newcommand{\cmark}{\ding{51}}
\newcommand{\xmark}{\ding{55}}
\usepackage{tabularx}
\usepackage{threeparttable} 
\usepackage[style=numeric, sorting=none, backend=biber]{biblatex} 
\usepackage{chngcntr}
\usepackage{listings}

\usepackage{xcolor}
\usepackage{amsmath,amssymb}
\title{RxnBench: A Multimodal Benchmark for Evaluating Large Language Models on
Chemical Reaction Understanding from Scientific Literature}

\author{
Hanzheng Li\textsuperscript{1,2}\thanks{These authors contributed equally to this work.} \and
Xi Fang\textsuperscript{1}\footnotemark[1] \and
Yixuan Li\textsuperscript{3}\footnotemark[1] \and 
Chaozheng Huang\textsuperscript{1} \and
Junjie Wang\textsuperscript{1,4} \and \\
Xi Wang\textsuperscript{5} \and
Hongzhe Bai\textsuperscript{6} \and
Bojun Hao\textsuperscript{7} \and
Shenyu Lin\textsuperscript{2} \and
Huiqi Liang\textsuperscript{8} \and \\
Linfeng Zhang\textsuperscript{1}\thanks{Corresponding authors.}  \and
Guolin Ke\textsuperscript{1}\footnotemark[2] \\
\affiliations
\textsuperscript{1}DP Technology \quad
\textsuperscript{2}Shanghai Jiao Tong University \quad
\textsuperscript{3}Tsinghua University \quad
\textsuperscript{4}Peking University\\
\textsuperscript{5}New York University \quad
\textsuperscript{6}Fudan University \quad
\textsuperscript{7}Xiamen University \quad
\textsuperscript{8}ShanghaiTech University
\emails \small
\{lihz@sioc.ac.cn,
fangxi@dp.tech,
l-yx21@mails.tsinghua.edu.cn,
huangchaozheng@dp.tech,
junjie-wang@pku.edu.cn,\\
xw3763@nyu.edu,
23300220029@m.fudan.edu.cn, 37420232204643@stu.xmu.edu.cn,
linshenyu@sjtu.edu.cn, \\lianghq2023@shanghaitech.edu.cn,
zhanglf@dp.tech,
kegl@dp.tech\}
}

\begin{document}

\maketitle

\thispagestyle{plain}



\begin{abstract}
The integration of Multimodal Large Language Models (MLLMs) into chemistry
promises to revolutionize scientific discovery, yet their ability to comprehend
the dense, graphical language of reactions within authentic literature remains
underexplored. Here, we introduce \textbf{RxnBench}, a multi-tiered benchmark
designed to rigorously evaluate MLLMs on chemical reaction understanding from
scientific PDFs. RxnBench comprises two tasks: Single-Figure QA (SF-QA), which
tests fine-grained visual perception and mechanistic reasoning using 1,525
questions derived from 305 curated reaction schemes, and Full-Document QA
(FD-QA), which challenges models to synthesize information from 108 articles,
requiring cross-modal integration of text, schemes, and tables. Our evaluation
of MLLMs reveals a critical capability gap: while models excel at extracting
explicit text, they struggle with deep chemical logic and precise structural
recognition. Notably, models with inference-time reasoning significantly
outperform standard architectures, yet none achieve 50\% accuracy on FD-QA.
These findings underscore the urgent need for domain-specific visual encoders
and stronger reasoning engines to advance autonomous AI chemists.
\end{abstract}

\section{Introduction}\label{introduction}

\begin{figure*}[t]  
  \centering
  \includegraphics[width=0.75\textwidth]{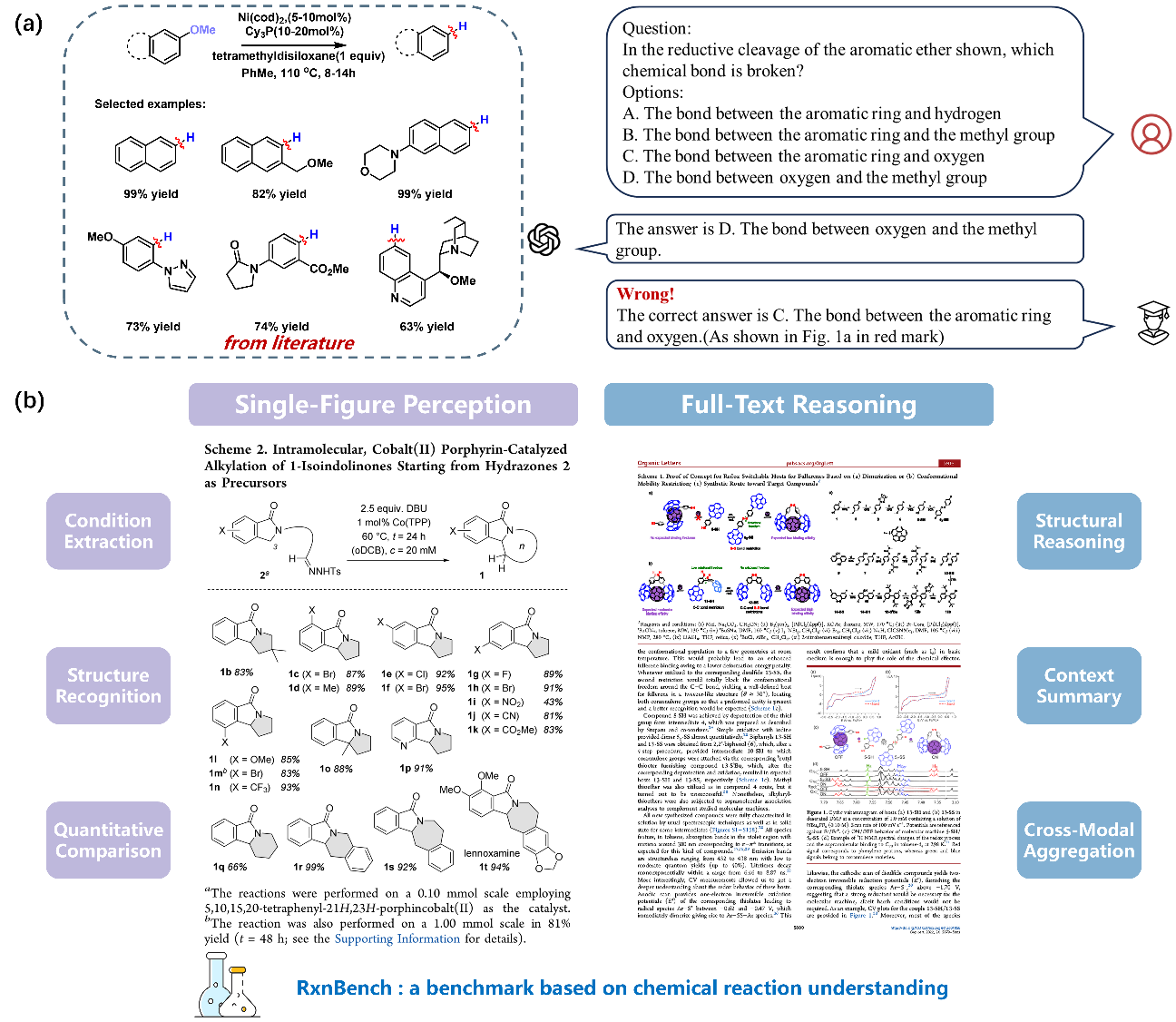} 
  \caption{\textbf{The RxnBench framework for advancing chemical
reaction understanding. (a) Challenges in reaction process understanding}.
Interpreting reaction schemes transcends simple object detection, requiring
models to infer dynamic roles---reactants, intermediates, and products---within
complex chemical transformations. \textbf{(b) Two-tiered evaluation hierarchy}.
RxnBench simulates a chemist's cognitive workflow through: (i) Single-Figure QA
(SF-QA) for granular visual perception and local mechanistic reasoning; and (ii)
Full-Document QA (FD-QA) for cross-modal synthesis and reasoning within
unstructured PDF documents.}
  \label{figure1}
\end{figure*}

The past decade has witnessed a paradigm shift in scientific discovery driven by
Artificial Intelligence (AI), particularly in chemistry, the central science,
where data-driven approaches are reshaping traditional research
workflows~\cite{Wang2023}. From early Quantitative
Structure-Activity Relationship (QSAR) models~\cite{Cherkasov2014}
to modern deep learning-based molecular
generation~\cite{GomezBombarelli2018} and property
prediction~\cite{Yang2019}, AI has demonstrated potential that
often transcends human intuition. Notably, the breakthrough in protein structure
prediction achieved by geometric deep learning models like
AlphaFold~\cite{Jumper2021} has significantly inspired the
chemical community to explore AI applications in small molecule drug
discovery~\cite{Vamathevan2019}, materials
design~\cite{Merchant2023}, and complex organic retrosynthesis
planning~\cite{Segler2018,Coley2019}.

In recent years, the emergence of Large Language Models (LLMs) has introduced a
new variable to chemical research. Unlike specialized models focused on single
tasks, LLMs exhibit surprising general reasoning capabilities and knowledge
integration skills~\cite{OpenAI2023}. Through instruction tuning
or agent-based frameworks (such as ChemCrow~\cite{Bran2023}),
LLMs have been proven capable of performing literature retrieval, experimental
planning, and even controlling automated laboratory
equipment~\cite{Boiko2023}. However, most existing chemical
large models rely heavily on linearized text representations (such as
SMILES~\cite{Weininger1988} or
SELFIES~\cite{Krenn2022}) to process chemical information.
Although the textual modality is a significant carrier of knowledge, the core
knowledge in chemical literature is often highly condensed within the
unstructured visual modality, limiting the complete acquisition and
understanding of chemical knowledge by text-only LLMs.

There are unique challenges in chemical literature understanding. Achieving
machine reading of chemical literature faces multiple challenges not encountered
in general document understanding tasks: First, the chemical reaction scheme is
a highly condensed domain-specific graphical language. Unlike natural images, a
reaction scheme is a complex semantic network composed of molecular structures,
arrow symbols, reaction conditions (temperature, solvents, catalysts, etc.), and
text labels. As shown in \textbf{Figure 1a}, understanding a reaction scheme
requires not only recognizing discrete molecular structures but also inferring
their roles in a dynamic chemical process (reactants, products, intermediates,
or transition states). This type of ``process understanding'' is far more
complex than simple object detection. Second, the unstructured nature of PDF
documents poses significant engineering and cognitive hurdles. Scientific
literature is primarily disseminated in PDF format---a format designed for human
visual consumption rather than machine parsing---often referred to as a
``graveyard'' for data~\cite{Tkaczyk2015}. In chemical papers,
reaction scheme images are frequently spatially separated from their textual
descriptions, requiring models to handle cross-page layout dependencies. Third,
the strong association and complementarity of multimodal information (Strong
Multimodality). Chemical knowledge rarely exists in a single modality in
isolation. For instance, a code in the main text (e.g., ``Compound 3a'') must be
aligned with the structure in an illustration (e.g., ``Scheme 1''), while
specific yields or enantiomeric excess (ee) values might be hidden in a
supplementary table (e.g., ``Table S2''). Models require the ability to perform
joint reasoning across text, images, and
tables~\cite{Yin2024}. Finally, the extreme requirement for
precision. Chemistry is an exact science. In natural image captioning,
misidentifying a ``dog'' as a ``cat'' is an error; however, in chemistry, a mere
change in the direction of a chemical bond (wedge vs. dash) implies a reversal
of stereochemistry, which in medicinal chemistry can mean the difference between
a cure and a poison (e.g., the thalidomide tragedy). Existing multimodal models
suffer from severe ``hallucination'' issues~\cite{Nori2023},
tending to generate structures that appear plausible but are chemically
incorrect.

As for the existing benchmarks for chemistry, we categorize them into two types
based on input modality: single-modality benchmarks, which present problems
purely in textual form, and multimodality benchmarks, which incorporate both
textual and visual inputs, such as molecular structures or crystallographic
images. Representative single-modal benchmarks include
ChemBench~\cite{Mirza2024},
ChemLLMBench~\cite{Guo2023},
MMLU~\cite{Hendrycks2020},
SciBench~\cite{Wang2023SciBench},
SciCode~\cite{Tian2024},
JEEBench~\cite{Arora2023}, and
OlympicArena~\cite{He2024}. While these provide robust
evaluation for textual reasoning, they inherently lack the ability to assess
visual chemical literacy. Recently, MOLERR2FIX ~\cite{wu2025molerr2fix} introduced a more granular evaluation paradigm by tasking models with detecting, localizing, and correcting fine-grained chemical errors within molecular descriptions. Conversely, multimodal initiatives such as
ScienceQA~\cite{Saikh2022},
OlympiadBench~\cite{He2024},
ChemTable~\cite{Zhou2025},
Mol-Puzzle~\cite{Guo2024}, and
ScemQA~\cite{Liang2024} have integrated visual data to assess
broader scientific literacy. Yet, a critical gap remains: these benchmarks
primarily target static concept recognition or elementary interpretation, often
treating molecular graphs as isolated objects rather than components of a
dynamic reaction process. They generally lack the granularity required to
evaluate the precise structural and quantitative reasoning---such as evaluating
reaction yields of different products, or verifying stereochemical
fidelity---that is central to organic synthesis. Furthermore, none of these
datasets challenge models to synthesize information from the full, unstructured
context of reaction-containing PDF literatures, a capability that fundamentally
distinguishes our work from existing concept-centric evaluations. This
limitation motivates the development of RxnBench, which aims to address these
gaps through broader coverage of reaction types and a distinct focus on reaction
perception and full-document reasoning.

\emph{\textbf{Our Contributions}}. To bridge this gap, we introduce RxnBench, a
benchmark designed specifically to evaluate the chemical reaction literature
understanding capabilities of Multimodal Large Language Models (MLLMs). The core
design philosophy of RxnBench is to simulate the authentic cognitive process of
a chemist reading literature, comprising tasks at two levels (\textbf{Figure
1b}):

\begin{enumerate}
\def\labelenumi{(\arabic{enumi})}
\item
  Single-Figure QA (SF-QA): Focuses on the ``perception'' level. We have
  constructed a dataset containing a high diversity of reaction types, requiring
  models to precisely extract reactant structures (SMILES), reaction conditions,
  yields, and stereochemical information from reaction scheme images. This aims
  to evaluate the model\textquotesingle s parsing precision of the chemical
  graphical language.
\item
  Full-Document QA (FD-QA): Focuses on the ``reasoning'' level. This is the
  first chemical benchmark requiring models to answer questions within the
  context of complete PDF literature. Models must integrate main text
  descriptions, reaction mechanism diagrams, and experimental data tables to
  perform cross-modal information retrieval and comprehensive reasoning.
\end{enumerate}

RxnBench aims to provide the community with a standardized evaluation tool, not
only revealing the capability boundaries of current State-of-the-Art (SOTA)
scientific multimodal models but also pointing the way toward the evolution of
more precise and reasoning-capable ``AI Chemists''.

\section{Construction of RxnBench}\label{construction-of-rxnbench}

\begin{figure*}[t]  
  \centering
  \includegraphics[width=0.75\textwidth]{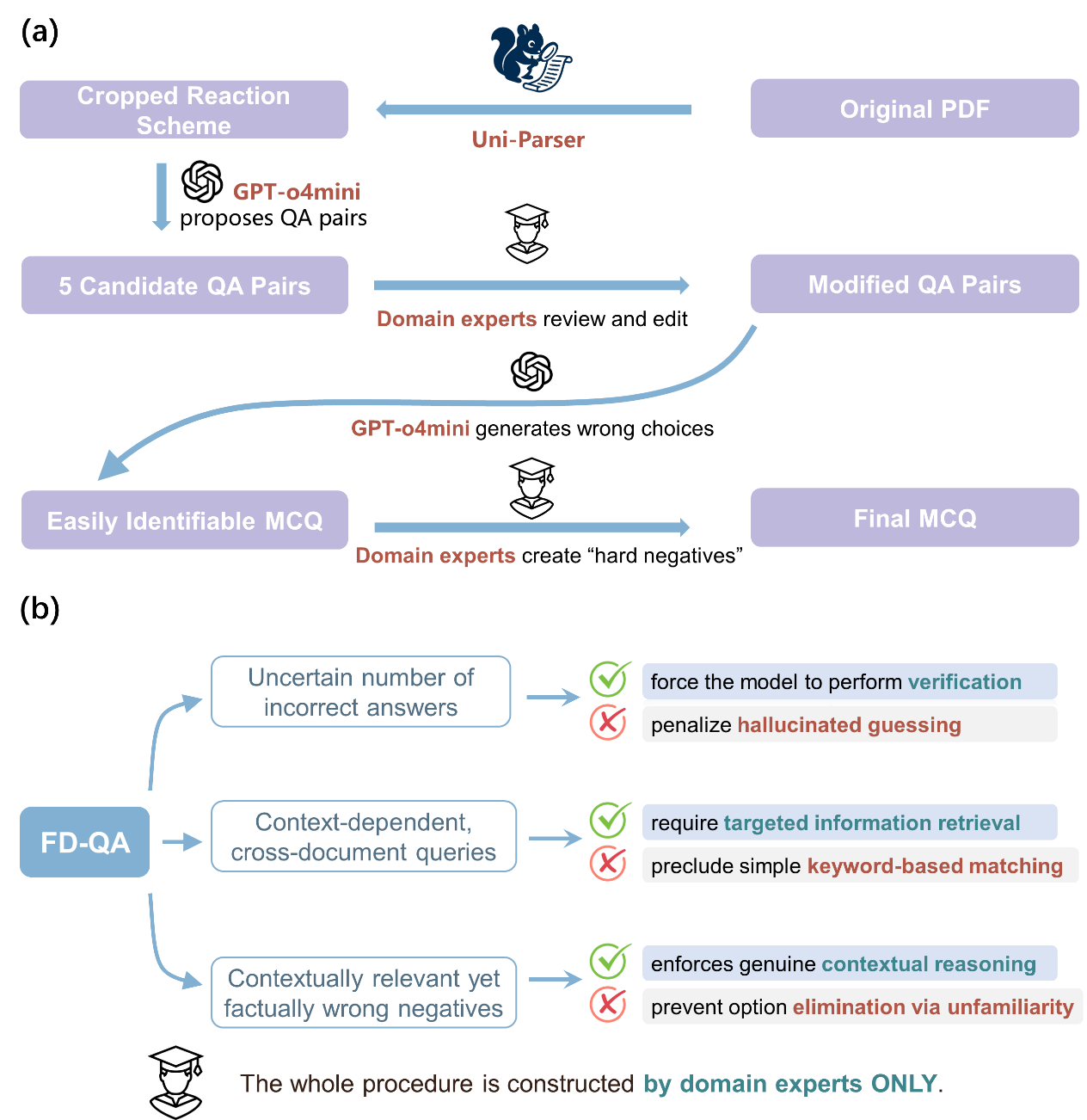} 
  \caption{\textbf{Construction methodologies and evaluation logic of RxnBench}.
  \textbf{(a) SF-QA construction pipeline}. A systematic workflow integrating
  automated extraction, human-in-the-loop refinement, and adversarial editing to
  ensure high-quality, chemically rigorous question-answer pairs. \textbf{(b)
  FD-QA evaluation hierarchy}. A comprehensive framework simulating a chemist's
  real-world workflow, requiring multi-page synthesis and cross-modal reasoning
  across text, schemes, and tables within complete scientific articles.}
  \label{figure2}
\end{figure*}

To ensure RxnBench serves as a rigorous and representative testbed for modern
chemical reasoning, we implemented a meticulous data curation and annotation
pipeline, prioritizing high-quality sources and domain-expert verification.

\subsection{Data Curation and Source Selection.}

We constructed our corpus from high-impact scientific articles published within
the last five years. To guarantee the diversity and complexity of the reaction
data, we selected Open Access (OA) papers from eight prestigious journals that
define the frontier of chemical research. These include top-tier
multidisciplinary journals and leading specialized journals in organic and
catalytic chemistry (\emph{Nature Chemistry, Nature Communications, JACS Au,
Nature Synthesis, Journal of the American Chemical Society, Angewandte Chemie
International Edition}, \emph{ACS Catalysis,} and \emph{Organic Letters}). Our
selection process focused on articles centered around organic reaction
methodology and total synthesis. We filtered candidates to ensure every selected
document contained:

\begin{enumerate}
\def\labelenumi{(\arabic{enumi})}
\item
  Rich Graphical Information: At least one comprehensive reaction scheme
  depicting reactants, reagents, conditions, and products with clear
  stereochemical notation (if have).
\item
  Complex Text-Image Interplay: Full-text descriptions that necessitate
  cross-referencing with the figures (e.g., discussion of reaction mechanisms or
  optimization tables).
\item
  Chemical Diversity: A broad coverage of reaction classes, ranging from classic
  cross-couplings to contemporary photoredox catalysis and C--H activation,
  ensuring the benchmark is not biased toward simple textbook reactions.
\end{enumerate}

Given the subtle nuances of organic chemistry---where a single stereocenter
inversion or a misinterpreted abbreviation can invalidate a
reaction---crowdsourcing platforms were deemed unsuitable for this task.
Instead, we recruited a team of domain experts to perform the annotation. The
annotation team consisted of doctoral candidates (Ph.D. students) specializing
in organic chemistry. All annotators possessed extensive experience in reading
synthetic literature and were proficient in using chemical drawing software and
interpreting mechanism of organic reactions. Furthermore, they were skilled in
Optical Chemical Structure Recognition (OCSR) workflows, to convert graphical
structures directly into Extended SMILES
(E-SMILES)~\cite{Fang2024}, facilitating precise structural
editing and verification.

\subsection{Construction of Single-Figure QA (SF-QA)}

The construction of the SF-QA dataset followed a systematic workflow designed to
ensure both scalability and depth (\textbf{Figure 2a}).

Figure Extraction and Segmentation is the first step. We utilized
Uni-Parser~\cite{Fang2025}, a specialized layout analysis tool
for scientific documents, to process the source PDFs. Reaction schemes were
automatically detected, cropped, and saved as high-resolution single images.
Crucially, this process was designed to identify and include the
scheme\textquotesingle s associated titles and labels (subscripts) within the
cropped area, thereby ensuring the completeness of the visual information. This
approach guarantees that the visual inputs for the benchmark contain all
necessary internal context while being isolated from the surrounding main text,
focusing the task strictly on visual chemical perception and preventing
information leakage from the body text.

The second step involves Hybrid Question Generation, where we adopted a
``Human-in-the-Loop'' strategy to produce high-quality Question-Answer (QA)
pairs. Initially, OpenAI o4-mini~\cite{Hurst2024} was seeded with
specific chemical templates to propose a set of five candidate QA pairs per
image. (The prompt was documented in \textbf{Supporting Information S4}). Domain
experts then acted as ``exam setters'', rigorously reviewing, modifying, or
completely rewriting these questions to ensure chemical relevance. These
questions were strictly categorized into two tiers to assess different levels of
cognitive capability. The first tier, \emph{\textbf{Descriptive Questions}},
focuses on information extraction, assessing the model\textquotesingle s ability
to accurately ``read'' the diagram. This encompasses tasks such as identifying
reaction titles, enumerating reactants and products, and extracting precise
reaction parameters like temperature, reaction time, and specific activation
methods (e.g., \emph{microwave irradiation, ball milling, or photoredox
conditions}). It also involves distinguishing the roles of various species, such
as differentiating a metal catalyst from its ligand. The second tier,
\emph{\textbf{Reasoning Questions}}, requires synthesizing visual information
with chemical knowledge to derive conclusions. This ranges from simple
reasoning, such as comparing yields and enantioselectivity across substrates or
performing Markush enumeration, to complex reasoning tasks like mechanism
analysis, identifying bond cleavage sites, and tracing catalytic cycles.
Advanced inquiries further test the model\textquotesingle s ability to analyze
structure-activity relationships (SAR) based on electronic effects and perform
retrosynthetic planning.

Following the validation of correct answers, we proceeded to construct a
rigorous multiple-choice evaluation framework through a two-stage process
involving automated generation and expert refinement. Initially, OpenAI o4-mini was
utilized to propose three distractor options for each validated QA pair.
However, recognizing that raw LLM-generated distractors are often plagued by
hallucinations or formatting artifacts that make them easily identifiable, our
domain experts performed a final round of ``Adversarial Editing'' to create
``Hard Negatives''. This step was critical to ensuring that the benchmark tests
genuine chemical understanding rather than the model\textquotesingle s ability
to exploit test-taking heuristics.

The adversarial refinement process adhered to the following principles. First,
experts minimized hallucinations, removing chemically impossible valencies or
non-existent functional groups to ensure all options represented valid chemical
entities. To prevent ``gaming'' via pattern matching, we standardized the format
and length of all options, minimizing ``length bias'' where correct answers
might otherwise stand out due to excessive detail. Crucially, the distractors
were designed to enforce visual dependency; options were constructed to be
chemically plausible yet mutually exclusive based on the visual evidence (e.g.,
offering both \emph{E-} and \emph{Z-} isomers), compelling the model to analyze
the specific reaction scheme rather than relying on general chemical knowledge.
Finally, we maximized chemical plausibility by mimicking common human
errors---such as including enantiomers, diastereomers, or regioisomers---thereby
ensuring that high performance on RxnBench reflects deep chemical insight.

\subsection{Construction of Full-Document QA (FD-QA)}

The Full-Document QA (FD-QA) task represents the pinnacle of the RxnBench
evaluation hierarchy, designed to test a model\textquotesingle s ability to
synthesize information from complete scientific articles---mimicking the daily
workflow of a research chemist(\textbf{Figure 2b}). Based on 108 representative
articles, we constructed a dataset that demands rigorous cross-modal reasoning
rather than simple keyword matching. Unlike the single-figure task, FD-QA is
formulated as a variable-response multiple-choice problem where each question
offers four substantive options (A-D) containing zero to four correct answers.
Crucially, to penalize models that rely on ``hallucinated guessing'', we
introduced a ``None of the Above'' option (Option E), which must be selected if
no correct answer is present among the provided choices. This design forces the
model to perform verification rather than mere selection.

A distinguishing feature of FD-QA, compared to the predominantly text-based
options in SF-QA, is the diversity of option formats designed to comprehensively
evaluate visual-linguistic alignment. Options are presented in three distinct
modalities: pure text containing standard chemical descriptions or numerical
values; pure structure images (e.g., cropped from ChemDraw) that require the
model to directly perceive and compare visual graphs without intermediate
textual translation; and mixed text-image layouts where text descriptions are
interleaved with structural snippets, simulating the complex layout often found
in reaction scope tables.

Domain experts were instructed to design five diverse questions per article,
adhering strictly to the principle of ``Context-Dependent Reasoning''. Rather
than simply requiring the presence of multiple modalities, it ensures that
answering necessitates synthesizing information distributed across the
document\textquotesingle s full context---bridging textual narratives, graphical
schemes, and tabular data. These questions encompass four advanced capability
dimensions: multimodal information extraction, which involves retrieving and
structuring data like product lists and yields embedded within complex layouts;
contextual summarization, which requires synthesizing scattered details to
reconstruct experimental procedures; chemical logic and reasoning, for tasks
demanding scientific deduction such as explaining regioselectivity based on a
mechanistic discussion; and reference resolution, which tests the ability to
navigate the document to map specific textual entity mentions to their
corresponding structural representations in the schemes.

To ensure rigor, the distractors in FD-QA are designed as contextually relevant
but factually incorrect ``hard negatives.'' Annotators ensured that all
incorrect options were derived directly from the PDF content (e.g., using
conditions from a different entry in the same table) rather than being
fabricated, thereby preventing models from ruling out options simply based on
unfamiliarity. Furthermore, to rigorously evaluate fine-grained structural
reasoning, we introduced a specialized category of Image-based Distractors.
These options consist of chemically valid structures generated via professional
drawing software that are visually ``plausible but incorrect.'' They are
intentionally crafted to bear high similarity to the ground truth while
differing in subtle, easily overlooked details---such as the regio-position of
substituents, variations in carbon chain length, or specific stereochemical
configurations---forcing the model to perform precise graph-level structural
verification rather than relying on coarse visual features.

\section{Results and discussion}\label{results-and-discussion}

\subsection{Task Categorization and Distribution in SF-QA}

To provide a granular analysis of model capabilities, we developed a
comprehensive taxonomy for the Single-Figure QA (SF-QA) dataset, classifying the
1,525 curated questions into six distinct types based on the cognitive skills
required. The definitions and distribution of these task types are summarized in
\textbf{Figure 3}.

\begin{figure*}[h]  
  \centering
  \includegraphics[width=0.75\textwidth]{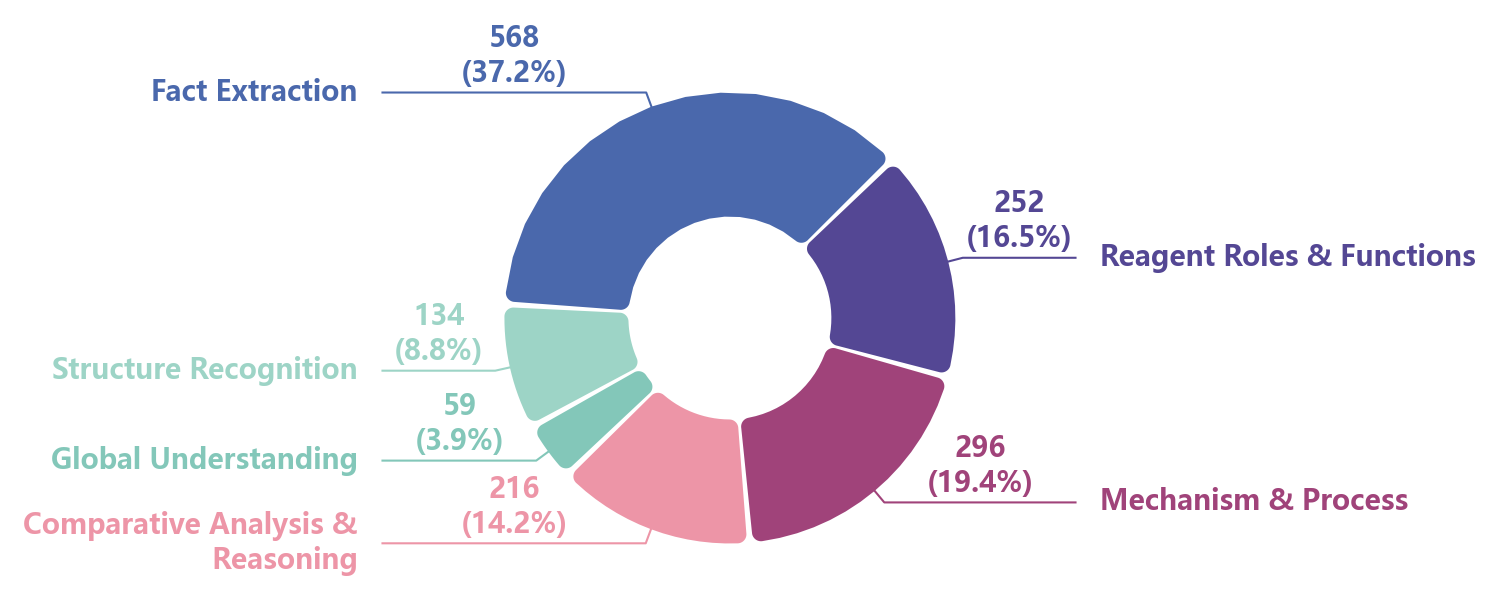} 
  \caption{\textbf{Distribution of Question Types in the SF-QA Dataset}. The
dataset covers 6 balanced types of tasks, ranging from fundamental information
extraction to advanced mechanistic reasoning and structural recognition.}
  \label{figure3}
\end{figure*}

\textbf{Fact Extraction} (37.2\%, 568 Questions): This foundational category
tests the model\textquotesingle s ability to directly retrieve explicit textual
or numerical information embedded within reaction schemes. Tasks include
identifying reaction titles, extracting yield values, or reading reaction times
and temperatures without requiring chemical interpretation. (e.g., ``\emph{What
is the title of this chemical reaction scheme?}'')

\textbf{Reagent Roles \& Functions} (16.5\%, 252 Questions): Moving beyond OCR,
this type requires the model to understand the chemical semantics of the text
labels. Models must distinguish between reagents, catalysts, solvents, and
additives, and correctly identify their functional roles (e.g., ``\emph{Which
molecule acts as the oxidant in this transformation?}'').

\textbf{Mechanism \& Process} (19.4\%, 296 Questions): This category assesses
the model's grasp of the dynamic nature of chemical reactions. Questions focus
on interpreting the reaction progression, including identifying key
intermediates, tracing catalytic cycles, and understanding the sequence of
mechanistic steps depicted in the scheme. (e.g., ``\emph{Which of the following
statements correctly describes the role of photocatalyst (PC) in the catalysis
cycle?}'')

\textbf{Comparative Analysis \& Reasoning} (14.2\%, 216 Questions): These
questions require the model to perform comparative evaluations or causal
reasoning. Typical tasks include predicting how changes in substrates (e.g.,
electronic effects of substituents) influence reaction outcomes like yield or
selectivity, often necessitating an analysis of provided scope tables. (e.g.,
``\emph{Compare the effects of different LED power and wavelength on the
reaction yield.}'')

\textbf{Global Understanding} (3.9\%, 59 Questions): Representing a higher-level
challenge, this type tests the comprehension of multi-step synthetic pathways.
Models must demonstrate an understanding of step-to-step coherence and the
overall synthetic design strategy. (e.g., ``\emph{Please describe the overall
transformation type and the key reaction processes of the electrochemical
reaction shown in the figure.}'')

\textbf{Structure Recognition} (8.8\%, 134 Questions): This task focuses on the
precise translation of visual molecular graphs into machine-readable strings
(SMILES or E-SMILES). Unlike general image captioning, this requires rigorous
graph-level parsing to capture exact connectivity, stereochemistry, and
functional group identity, serving as a direct evaluation of the
model\textquotesingle s ``chemical vision.'' (e.g., ``\emph{What is the E-SMILES
of the product containing a tert-butyl group?}'')

As illustrated in \textbf{Figure 3}, while Fact Extraction constitutes the
largest portion (37.2\%) to ensure a solid baseline for visual perception, the
majority of the benchmark (62.8\%) is dedicated to tasks requiring varying
degrees of chemical knowledge and reasoning. Notably, nearly one-fifth of the
questions are dedicated to mechanism understanding, highlighting
RxnBench\textquotesingle s focus on deep chemical logic rather than superficial
pattern matching.

\subsection{Overall Model Performance on SF-QA}

We evaluated a comprehensive suite of Multimodal Large Language Models (MLLMs),
ranging from proprietary giants to state-of-the-art open-weight models, using a
Zero-shot prompting strategy. To assess linguistic robustness, all models were
queried using both English (RxnBench-En) and Chinese (RxnBench-Zh) prompts. To
handle non-standard outputs where models typically generate verbose explanations
instead of a specific option, we employed a GPT-4o-based mapping strategy to
align free-text responses to the standard A--D options, ensuring a fair
comparison across instruction-following capabilities.

\begin{table*}[t]  
    \caption{Performance comparison of selected representative MLLMs on the SF-QA dataset.}
    {\small \textit{Models are ranked by Mean Score.}} \\ [1ex] 
    \begin{tabularx}{\textwidth}{l X X c c c}
        \hline
        Model & Think\textsuperscript{a} & Weight\textsuperscript{b} & RxnBench-En & RxnBench-Zh & Mean Score\\
        \hline
        Gemini-3-Flash-preview & \cmark  & - & \textbf{0.9593} & \textbf{0.9652} & \textbf{0.9623}\\
        Seed1.8-Think & \cmark  & - & 0.9325 & 0.9403 & 0.9364\\
        Gemini-3-Pro-preview & \cmark  & - & 0.9318 & 0.9403 & 0.9361\\
        GPT-5(high) & \cmark  & - & 0.9279 & 0.9246 & 0.9263\\
        Gemini-2.5-Pro & \cmark  & - & 0.9095 & 0.9423 & 0.9259\\
        Qwen3-VL-235B-A22B-Think & \cmark & Open & 0.9220 & 0.9134 & 0.9177\\
        Seed1.5-VL-Think & \cmark  & - & 0.9056 & 0.9161 & 0.9109\\
        InternVL3.5-241B-A28B & \cmark & Open & 0.9003 & 0.9062 & 0.9033\\
        Intern-S1 & \cmark & Open & 0.8938 & 0.8944 & 0.8941\\
        Seed1.5-VL & \xmark & - & 0.8518 & 0.8669 & 0.8594\\
        Qwen3-VL-235B-A22B-Instruct & \xmark & Open & 0.8492 & 0.8675 & 0.8584\\
        GPT-4o & \xmark & - & 0.7462 & 0.7436 & 0.7449\\
        Random & - & - & 0.2500 & 0.2500 & 0.2500\\
        \hline
    \end{tabularx}
\end{table*}

To comprehensively map the capability landscape, we evaluated a total of 41
Multimodal Large Language Models, spanning diverse architectures (from
lightweight to massive-scale) and access types (proprietary vs. open-source).
The overall accuracy results for selected representative models are summarized
in \textbf{Table 1} (see \textbf{Supplementary Table S1} for the complete data
of all evaluated models). Overall, the evaluation highlights a performance
hierarchy driven by model scale and inference-time reasoning, where proprietary
frontiers and ``thinking'' models lead, yet open-source alternatives are rapidly
narrowing the gap with robust bilingual consistency.

As shown in \textbf{Table 1}, the leaderboard is currently dominated by
next-generation proprietary models. Gemini-3-Flash-preview achieves the highest
mean score of 96.23\%, demonstrating exceptional capability in chemical vision
understanding. It is closely followed by Seed1.8-Think (93.64\%) and
Gemini-3-Pro-preview (93.61\%). These results suggest that the scaling laws of
model size and data quality continue to drive significant improvements in
scientific domain tasks.

A notable trend is the superior performance of models equipped with
inference-time reasoning capabilities (denoted as ``Think'' in \textbf{Table
1}). For instance, Compare with Qwen3-VL-235B-A22B-Instruct (85.84\%),
Qwen3-VL-235B-A22B-Think achieves a mean score of 91.77\%, effectively bridging
the gap between open-weight and top-tier proprietary models. It indicates that
allowing the model to generate internal chains of thought before outputting an
answer is particularly beneficial for the complex, multi-step reasoning required
in chemical reaction analysis.

While proprietary models lead the pack, open-source models have made remarkable
strides. The Qwen3-VL and InternVL series perform competitively, with the best
open models surpassing previous generation proprietary models like GPT-4o
(74.49\%) by a significant margin. This democratization of capability is
promising for the deployment of local AI chemists. Furthermore, most top-tier
models exhibit high consistency between English and Chinese benchmarks (e.g.,
Gemini-3-Flash-preview: 95.93\% EN vs 96.52\% ZH), reflecting strong
cross-lingual alignment in their pre-training, though a slight performance drop
is observed in some smaller models when processing Chinese prompts.

\subsection{Detailed Analysis by Question Type on SF-QA}

To dissect the fine-grained capabilities of current MLLMs, we decomposed the
SF-QA performance across the six defined task types. The results, detailed in
\textbf{Table 2}(see \textbf{Supplementary Table S2} for the complete data of
all evaluated models), reveal significant variance in model proficiency
depending on the cognitive depth required.

A distinct capability gap emerges as tasks increase in complexity. For
fundamental \textbf{Fact Extraction}, top-tier models demonstrate near-perfect
proficiency, with Gemini-3-Pro-preview achieving 96.48\% and
Gemini-3-Flash-preview reaching 96.13\%, indicating that basic information
retrieval is effectively solved. However, in complex reasoning tasks, a clear
divergence appears. Standard baselines like GPT-4o show limited capability
(scoring 79.73\% and 76.27\% respectively), whereas models equipped with
inference-time reasoning maintain exceptional robustness. For instance, GPT-5
(high) achieves a remarkable 96.61\% on \textbf{Global Understanding}, and
Gemini-3-Flash-preview achieves 97.64\% on \textbf{Mechanism \& Process}. This
performance contrast highlights a ``reasoning leap'', where internal chains of
thought successfully bridge the gap between perception and deep chemical logic.

Despite the reasoning advantage, a critical performance drop reappears in
\textbf{Structure Recognition} across the board. Even leading proprietary models
struggle to match their text-based performance in this category; for example,
Gemini-3-Pro-preview drops to 74.63\%, and GPT-4o falls to 52.24\%. Although
specific architectures like Qwen3-VL-235B-A22B-Think demonstrate a relative
advantage with 84.33\% and Gemini-3-Flash-preview reaches 90.30\%, the general
downward trend across most models suggests that reasoning alone cannot fully
compensate for the deficits in visual molecular encoding for all architectures
(\textbf{Figure 4}).

Collectively, these findings indicate that while current SOTA models have
mastered the textual ``reading'' of chemical literature, the frontiers of
accurate visual molecular perception and deep mechanistic ``understanding''
remain active challenges. The data suggests that future advancements must
prioritize domain-specific visual encoders to resolve the universal bottleneck
in structure recognition.

\begin{table*}[t]  
\begin{threeparttable}
    \caption{Detailed Model Performance selected across SF-QA Task Types}
    \begin{tabularx}{\textwidth}{@{} l X X X X X X X X @{}}
        \hline
        Model & Think\tnote{a} & Weight\tnote{b} & F.E.\tnote{c} & RR\&F.\tnote{d} & M\&P.\tnote{e} & CA\&R.\tnote{f} & G.U.\tnote{g} & S.R.\tnote{h}\\
        \hline
        Gemini-3-Flash-preview & \cmark  & - & 0.9613 & \textbf{0.9643} & \textbf{0.9764} & \textbf{0.9630} & 0.9492 & \textbf{0.9030}\\
        Seed1.8-Think & \cmark  & - & 0.9331 & 0.9484 & 0.9527 & 0.9444 & 0.9492 & 0.8284\\
        Gemini-3-Pro-preview & \cmark  & - & \textbf{0.9648} & 0.9246 & 0.9527 & 0.9398 & 0.9322 & 0.7463\\
        GPT-5(high) & \cmark  & - & 0.9313 & 0.9444 & 0.9527 & 0.9167 & \textbf{0.9661} & 0.8358\\
        Gemini-2.5-Pro & \cmark  & - & 0.9331 & 0.9246 & 0.9459 & 0.9491 & 0.9322 & 0.6343\\
        Qwen3-VL-235B-A22B-Think & \cmark & Open & 0.9190 & 0.9405 & 0.9459 & 0.9213 & 0.9322 & 0.8433\\
        Seed1.5-VL-Think & \cmark  & - & 0.8996 & 0.9365 & 0.9358 & 0.9074 & 0.9153 & 0.8060\\
        InternVL3.5-241B-A28B & \cmark & Open & 0.8944 & 0.9127 & 0.9291 & 0.9167 & 0.9153 & 0.8134\\
        Intern-S1 & \cmark & Open & 0.9014 & 0.9127 & 0.9223 & 0.9028 & 0.8814 & 0.7463\\
        Seed1.5-VL & \xmark & - & 0.9327 & 0.9127 & 0.9122 & 0.8472 & 0.8305 & 0.7015\\
        Qwen3-VL-235B-A22B-Instruct & \xmark & Open & 0.8204 & 0.8929 & 0.8986 & 0.8426 & 0.8814 & 0.7761\\
        GPT-4o & \xmark & - & 0.7359 & 0.8175 & 0.7973 & 0.7500 & 0.7627 & 0.5224\\
        Random & - & - & 0.2500 & 0.2500 & 0.2500 & 0.2500 & 0.2500 & 0.2500\\
        \hline
    \end{tabularx}
    \begin{tablenotes}
        \footnotesize
        \item[a] Think: indicates models with inference-time reasoning capabilities
        \item[b] Weight: distinguishes between Proprietary (Closed) and Open-weight models
        \item[c] F.E.: Fact Extraction
        \item[d] RR\&F.: Reagent Roles \& Functions
        \item[e] M\&P.: Mechanism \& Process
        \item[f] CA\&R.: Comparative Analysis \& Reasoning
        \item[g] G.U.: Global Understanding
        \item[h] S.R.: Structure Recognition
    \end{tablenotes}
\end{threeparttable}
\end{table*}

\begin{table*}[!h]  
    \begin{threeparttable}
        \caption{Detailed Model Performance selected across FD-QA Task Types}
        \label{table2}
        \begin{tabularx}{\textwidth}{@{} l X X X X X X X @{}}
            \hline
            Model & Think\tnote{a} & Weight\tnote{b} & Rxn.En.\tnote{c} & Rxn.Zh.\tnote{d} & C.R.\tnote{e} & S.R.\tnote{f} & O.S.\tnote{g}\\
            \hline
            Gemini-3-Flash-preview & \cmark  & - & \textbf{0.4574} & \textbf{0.4685} & 0.5530 & \textbf{0.4264} & \textbf{0.4630}\\
            Gemini-3-Pro-preview & \cmark  & - & 0.4167 & 0.4407 & 0.5000 & 0.3897 & 0.4287\\
            Gemini-2.5-Pro & \cmark  & - & 0.4148 & 0.4426 & \textbf{0.5682} & 0.3652 & 0.4287\\
            Gemini-2.5-Flash & \cmark  & - & 0.3519 & 0.3815 & 0.4394 & 0.3235 & 0.3667\\
            Seed1.8-Think & \cmark  & - & 0.3352 & 0.3889 & 0.4242 & 0.3064 & 0.3620\\
            GPT-5(high) & \cmark  & - & 0.3444 & 0.3463 & 0.4470 & 0.3113 & 0.3454\\
            Seed1.5-VL-Think & \cmark  & - & 0.3148 & 0.3574 & 0.3939 & 0.2892 & 0.3361\\
            Qwen3-VL-235B-A22B-Think & \cmark & Open & 0.3056 & 0.3056 & 0.4091 & 0.2721 & 0.3056\\
            Intern-S1 & \cmark & Open & 0.1963 & 0.1944 & 0.2652 & 0.1740 & 0.1954\\
            Seed1.5-VL-Instruct & \xmark & - & 0.1630 & 0.1741 & 0.2045 & 0.1495 & 0.1685\\
            GPT-4o & \xmark & - & 0.1426 & 0.1389 & 0.1515 & 0.1397 & 0.1407\\
            Qwen3-VL-235B-A22B-Instruct & \xmark & Open & 0.1241 & 0.1481 & 0.1742 & 0.1078 & 0.1361\\
            \hline
        \end{tabularx}
        \begin{tablenotes}
            \footnotesize
            \item[a] Think: indicates models with inference-time reasoning capabilities
            \item[b] Weight: distinguishes between Proprietary (Closed) and Open-weight models
            \item[c] Rxn.En.: RxnBench-Doc-En
            \item[d] Rxn.Zh.: RxnBench-Doc-Zh
            \item[e] C.R.: Context Reasoning (En)
            \item[f] S.R.: Structure Reasoning (En)
            \item[g] O.S.: Overall Score
        \end{tablenotes}
    \end{threeparttable}
\end{table*}

\subsection{Evaluation on Full-Document QA (FD-QA)}

Originated from 108 research articles, the FD-QA benchmark comprises 540
meticulously curated questions designed to assess document-level chemical
comprehension. These tasks are categorized into two distinct types:
\textbf{Context Reasoning} (24.4\%), which requires the integration of
information from multimodal sources such as reaction images, tables, and text to
answer questions; and \textbf{Structure Reasoning} (75.6\%), which focuses on
rigorous logical deductions related to molecular and Markush structures,
reaction components, and mechanistic inference.

\begin{figure}[H]
  \centering
  \includegraphics[width=0.45\textwidth]{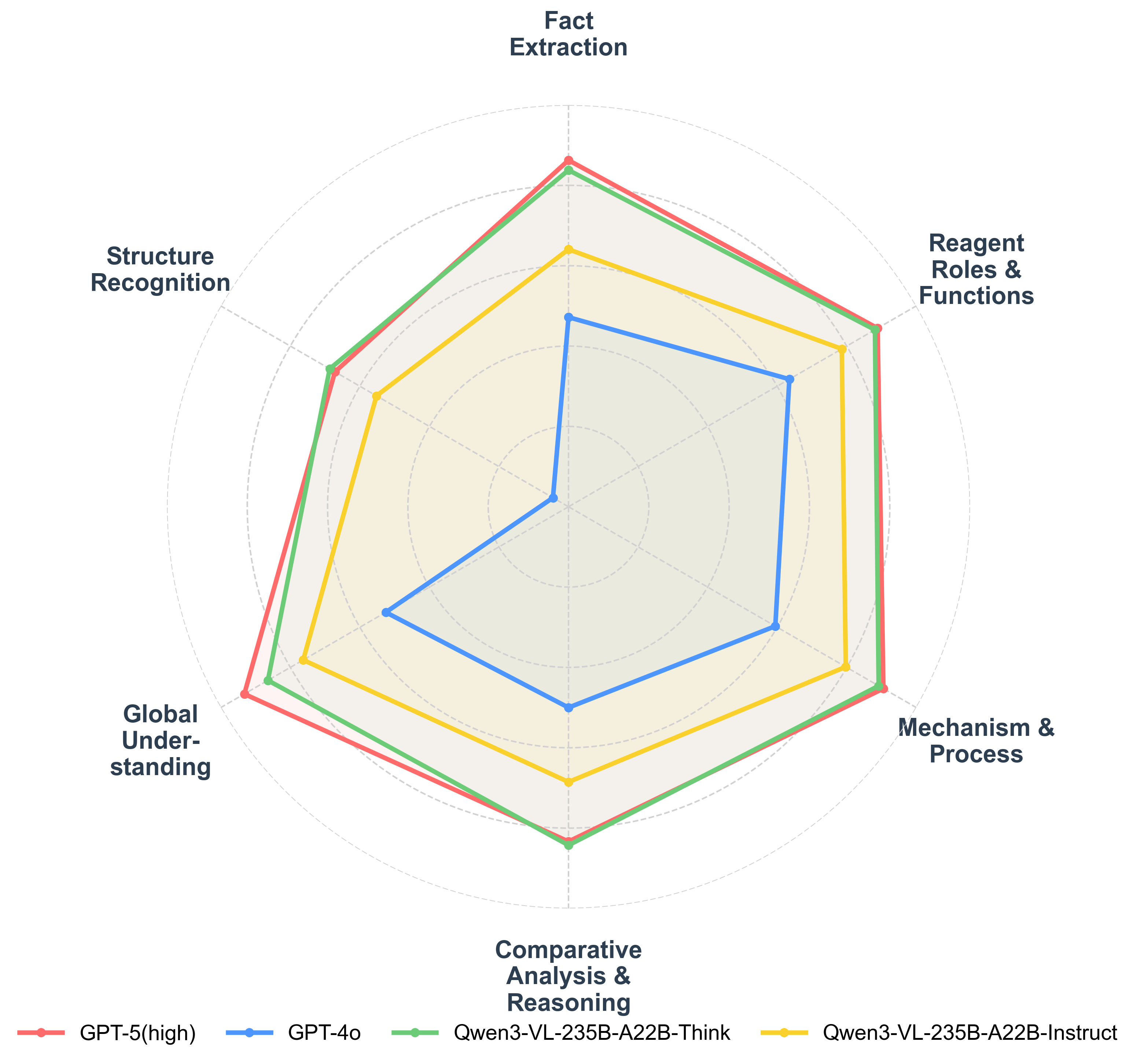} 
  \caption{\textbf{Comparative performance analysis of inference-time
reasoning (``Think'') versus standard instruction-tuned (``Instruct'') models
across diverse cognitive tasks.} The visualization highlights the ``reasoning
leap'' in complex mechanistic and synthetic tasks achieved by thinking models
(red/green lines) compared to standard baselines (blue/yellow lines), while also
illustrating the universal bottleneck in precise structure recognition.}
  \label{figure4}
\end{figure}

To simulate a realistic reading environment, we render each PDF document into a
sequence of high-resolution images at 144 dpi, serving as inputs to the MLLM.
The \textless answer\textgreater{} tag in the question acts as a placeholder
that is replaced by the corresponding image list, forming an interleaved
vision--language prompt. We then employ a GPT-4o-based extraction pipeline to
parse the model's raw output into a standard multiple-select format (A--E),
explicitly mapping statements such as ``None of the above'' to Option E. The
extracted answers are compared against the ground truth using an exact-match
criterion, where any missing or extraneous selections are counted as errors. We
report absolute accuracy for both the English and Chinese versions. Note that
cases where the model exceeds the allowable context length at inference time
result in missing scores, an issue observed in part of the Intern-S1 results.

We evaluated a total of 23 Multimodal Large Language Models. The results,
summarized in \textbf{Table 3}, reveal that FD-QA represents a significantly
higher difficulty tier compared to single-figure tasks. Note that due to space
constraints, \textbf{Table 3} highlights the performance of selected
representative models that define the current state-of-the-art or serve as key
baselines; the complete leaderboard featuring all 23 evaluated models is
provided in \textbf{Tabel S3} in \textbf{Supplementary Information}.

The evaluation exposes a stark performance chasm where models lacking
inference-time reasoning fail to surpass a 15\% accuracy threshold, whereas
``thinking'' models (particularly the Gemini family which leads with
\textasciitilde46\% accuracy) demonstrate superior capability in handling
long-context multimodal inputs, effectively validating the necessity of
Chain-of-Thought processing for complex document synthesis. Even so, the fact
that no model reaches 50\% accuracy underscores the substantial challenge of
synthesizing full scientific narratives compared to isolated perception tasks.

Crucially, a significant disparity exists between reasoning modalities. As shown
in the detailed breakdown, models consistently underperform on \textbf{Structure
Reasoning} compared to \textbf{Context Reasoning} (e.g., Gemini-2.5-Pro: 36.52\%
vs 56.82\%). This trend persists across almost all evaluated models
(\textbf{Figure 5}), indicating that while MLLMs are relatively adept at
retrieving and correlating textual or tabular information (context), their
ability to perform precise logical deductions based on visual molecular
structures remains a distinct bottleneck.

\begin{figure}[!h]
  \centering
  \includegraphics[width=0.45\textwidth]{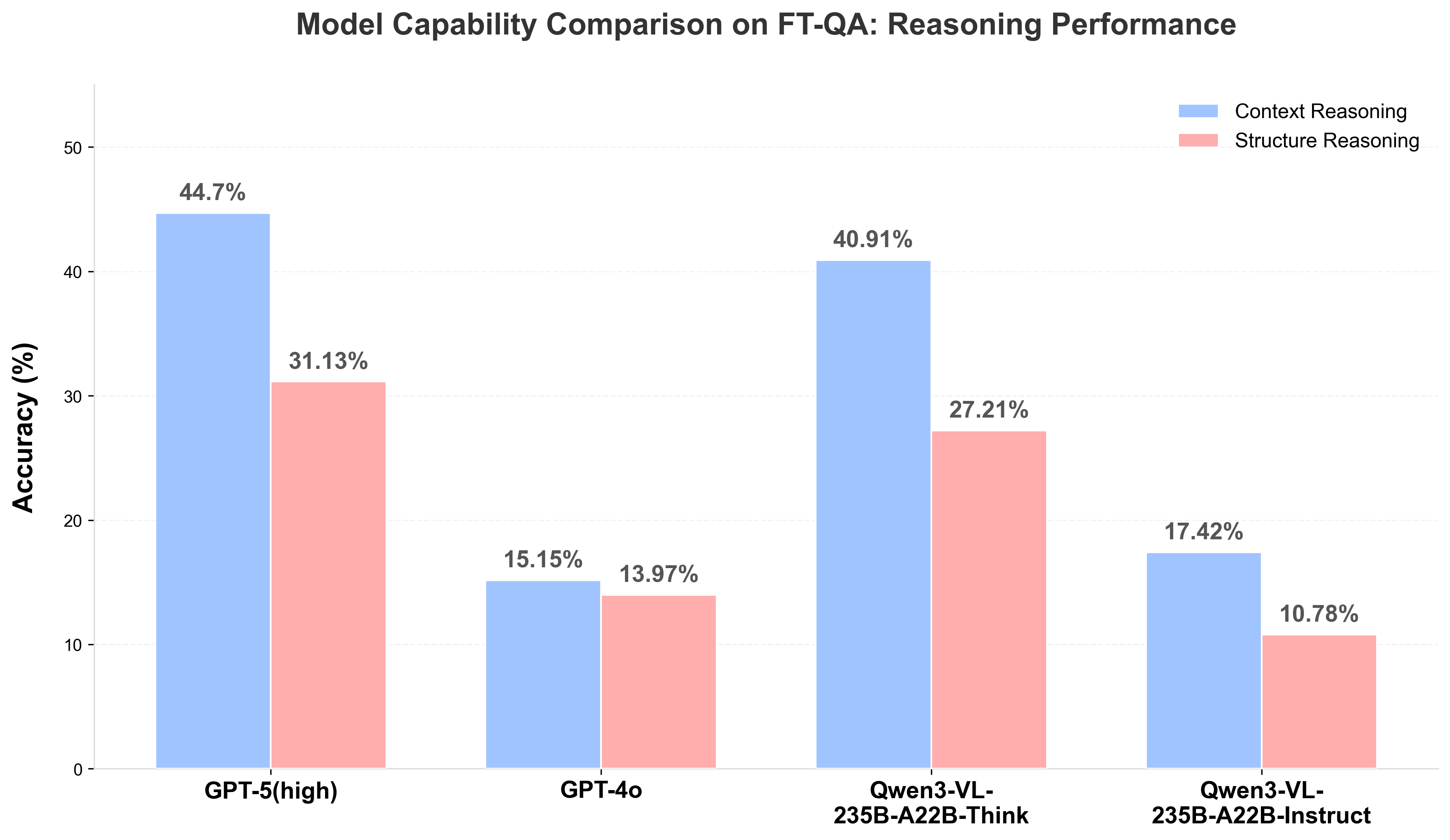} 
  \caption{\textbf{Comprehensive performance evaluation of MLLMs on FD-QA
benchmark.} The chart details the accuracy of various models across
\textbf{Context Reasoning} and \textbf{Structure Reasoning}. The results
highlight the substantial advantage of ``thinking'' models over standard
architectures in handling long-context scientific synthesis, while also
revealing a consistent performance lag in structure-dependent reasoning tasks.}
  \label{figure5}
\end{figure}

\section{Conclusion}\label{conclusion}

In this work, we introduced RxnBench, a pioneering multimodal benchmark
meticulously designed to evaluate the chemical reaction understanding
capabilities of Large Language Models within the context of authentic scientific
literature. By constructing a dual-tiered evaluation framework comprising
Single-Figure QA (SF-QA) and Full-Document QA (FD-QA), we have provided the
community with a rigorous standard to assess not just perceptual accuracy, but
the depth of chemical reasoning.

Our comprehensive evaluation reveals a distinct landscape where MLLMs have
mastered surface-level information extraction but face substantial bottlenecks
in deep chemical reasoning and precise structural recognition---critical
deficits that are partially mitigated by inference-time reasoning capabilities
yet remain acute when synthesizing complex, long-context scientific narratives.

RxnBench serves as a diagnostic tool that points the way forward. Future
research should prioritize the development of domain-specific visual encoders
capable of pixel-perfect molecular graph recognition and the integration of
external chemical tools (e.g., RDKit, calculator agents) to verify
hallucinations. Furthermore, moving beyond passive QA to active Agentic
Workflows---where models autonomously navigate, query, and verify literature
data---will be essential for realizing the vision of a truly autonomous AI
Chemist.

\section{Associated content}\label{associated-content}

\subsection{Data Availability Statement}

SF-QA Benchmark dataset is available at \url{https://huggingface.co/datasets/UniParser/RxnBench}

FD-QA Benchmark dataset is available at \url{https://huggingface.co/datasets/UniParser/RxnBench-Doc}. 

Code is available at \url{https://github.com/uni-parser/RxnBench}.

\subsection{Acknowledgements}

This research was supported by the New Generation Artificial Intelligence-National Science and Technology Major Project 2025ZD0121905.

We thank the students from Shanghai Jiao Tong University who contributed to the establishment of RxnBench (No particular order): Jing Guan, Jingli He, Panpan Li, Yifan Song, Tiandong Guan, Guanlin Li, Jiacheng Zhang, Qinyi Li, Shaoyi Dan, Lefeng Dong, Xintai Zhang, Xuezhen Kou, Jie Zhao, Yicong Luo, and Haoyang Wang.













\printbibliography

\clearpage






\setcounter{table}{0}
\renewcommand{\tablename}{}
\renewcommand{\thetable}{Table S\arabic{table}}


\begin{table*}[!hb]
\caption{Performance comparison of representative MLLMs on the SF-QA.}
\begin{tabularx}{\textwidth}{l X X c c c}
\hline
Model & Think & Weight & RxnBench-En & RxnBench-Zh & Mean Score\\
\hline
Gemini-3-Flash-preview & \cmark & - & \textbf{0.9593} & \textbf{0.9652} & \textbf{0.9623}\\
Seed1.8-Think & \cmark & - & 0.9325 & 0.9403 & 0.9364\\
Gemini-3-Pro-preview & \cmark & - & 0.9318 & 0.9403 & 0.9361\\
GPT-5(high) & \cmark & - & 0.9279 & 0.9246 & 0.9263\\
Gemini-2.5-Pro & \cmark & - & 0.9095 & 0.9423 & 0.9259\\
GPT-5.1(high) & \cmark & - & 0.9213 & 0.9220 & 0.9216\\
GPT-5(medium) & \cmark & - & 0.9207 & 0.9226 & 0.9216\\
Qwen3-VL-235B-A22B-Think & \cmark & Open & 0.9220 & 0.9134 & 0.9177\\
Qwen3-VL-32B-Think & \cmark & Open & 0.9128 & 0.9161 & 0.9144\\
GPT-5.1(medium) & \cmark & - & 0.9108 & 0.9141 & 0.9125\\
GPT-5-mini & \cmark & - & 0.9108 & 0.9128 & 0.9118\\
Seed1.5-VL-Think & \cmark & - & 0.9056 & 0.9161 & 0.9109\\
o3 & \cmark & - & 0.9056 & 0.9115 & 0.9086\\
o4-mini & \cmark & - & 0.9062 & 0.9075 & 0.9069\\
InternVL3.5-241B-A28B & \cmark & Open & 0.9003 & 0.9062 & 0.9033\\
Intern-S1 & \cmark & Open & 0.8938 & 0.8944 & 0.8941\\
Qwen3-VL-30B-A3B-Think & \cmark & Open & 0.8689 & 0.8590 & 0.8689\\
Qwen3-VL-Plus & \xmark & - & 0.8551 & 0.8656 & 0.8604\\
Qwen3-VL-8B-Think & \cmark & Open & 0.8636 & 0.8564 & 0.8600\\
Seed1.5-VL & \xmark & - & 0.8518 & 0.8669 & 0.8594\\
Qwen3-VL-235B-A22B-Instruct & \xmark & Open & 0.8492 & 0.8675 & 0.8584\\
InternVL3-78b & \xmark & Open & 0.8531 & 0.8308 & 0.8420\\
Qwen3-VL-4B-Think & \cmark & Open & 0.8577 & 0.8256 & 0.8416\\
Intern-S1-mini & \cmark & Open & 0.8521 & 0.8282 & 0.8402\\
GLM-4.1V-9B-Thinking & \cmark & Open & 0.8392 & 0.8341 & 0.8367\\
Qwen3-VL-32B-Instruct & \xmark & Open & 0.8315 & 0.8407 & 0.8361\\
Qwen2.5-VL-72B & \xmark & Open & 0.8341 & 0.8308 & 0.8325\\
Qwen2.5-VL-Max & \xmark & - & 0.8192 & 0.8262 & 0.8227\\
GPT-5-nano & \cmark & - & 0.7980 & 0.7941 & 0.7961\\
Qwen2.5-VL-32B & \xmark & Open & 0.7980 & 0.7908 & 0.7944\\
Gemini-2.5-Flash & \cmark & - & 0.6925 & 0.8557 & 0.7741\\
Qwen3-VL-8B-Instruct & \xmark & Open & 0.7548 & 0.7495 & 0.7521\\
Qwen3-VL-30B-A3B-Instruct & \xmark & Open & 0.7456 & 0.7436 & 0.7456\\
GPT-4o & \xmark & - & 0.7462 & 0.7436 & 0.7449\\
Qwen2.5-VL-7B & \xmark & Open & 0.7082 & 0.7233 & 0.7158\\
Qwen3-VL-4B-Instruct & \xmark & Open & 0.7023 & 0.7023 & 0.7023\\
Qwen3-VL-2B-Think & \cmark & Open & 0.6780 & 0.6708 & 0.6744\\
Qwen2.5-VL-3B & \xmark & Open & 0.6748 & 0.6643 & 0.6696\\
GPT-4o mini & \xmark & - & 0.6636 & 0.6066 & 0.6351\\
Qwen3-VL-2B-Instruct & \xmark & Open & 0.5711 & 0.5928 & 0.5820\\
Choose the longest answer & - & - & 0.4262 & 0.4525 & 0.4394\\
Deepseek-VL2 & \xmark & Open & 0.4426 & 0.4216 & 0.4321\\
Random & - & - & 0.2500 & 0.2500 & 0.2500\\
\hline
\end{tabularx}
\end{table*}

\begin{table*}
\begin{threeparttable}
  \caption{Performance of evaluated models across SF-QA Task Types}
  \begin{tabularx}{\textwidth}{@{} l X X X X X X X X @{}}
  \hline
  Model & Think\tnote{a} & Weight\tnote{b} & F.E.\tnote{c} & RR\&F.\tnote{d} & M\&P.\tnote{e} & CA\&R.\tnote{f} & G.U.\tnote{g} & S.R.\tnote{h}\\
  \hline
  Gemini-3-Flash-preview & \cmark & - & 0.9613 & \textbf{0.9643} & \textbf{0.9764} & \textbf{0.9630} & 0.9492 & \textbf{0.9030}\\
  Seed1.8-Think & \cmark & - & 0.9331 & 0.9484 & 0.9527 & 0.9444 & 0.9492 & 0.8284\\
  Gemini-3-Pro-preview & \cmark & - & \textbf{0.9648} & 0.9246 & 0.9527 & 0.9398 & 0.9322 & 0.7463\\
  GPT-5(high) & \cmark & - & 0.9313 & 0.9444 & 0.9527 & 0.9167 & \textbf{0.9661} & 0.8358\\
  Gemini-2.5-Pro & \cmark & - & 0.9331 & 0.9246 & 0.9459 & 0.9491 & 0.9322 & 0.6343\\
  GPT-5.1(high) & \cmark & - & 0.9243 & 0.9524 & 0.9426 & 0.9167 & 0.9661 & 0.7910\\
  GPT-5(medium) & \cmark & - & 0.9349 & 0.9325 & 0.9493 & 0.9167 & 0.9492 & 0.7761\\
  Qwen3-VL-235B-A22B-Think & \cmark & Open & 0.9190 & 0.9405 & 0.9459 & 0.9213 & 0.9322 & 0.8433\\
  Qwen3-VL-32B-Think & \cmark & Open & 0.9296 & 0.9405 & 0.9426 & 0.9259 & 0.9153 & 0.7015\\
  GPT-5.1(medium) & \cmark & - & 0.9243 & 0.9365 & 0.9426 & 0.9167 & 0.9492 & 0.7090\\
  GPT-5-mini & \cmark & - & 0.9225 & 0.9325 & 0.9257 & 0.9259 & 0.9831 & 0.7388\\
  Seed1.5-VL-Think & \cmark & - & 0.8996 & 0.9365 & 0.9358 & 0.9074 & 0.9153 & 0.8060\\
  o3 & \cmark & - & 0.9313 & 0.9325 & 0.9223 & 0.8981 & 0.9492 & 0.7090\\
  o4-mini & \cmark & - & 0.6391 & 0.7302 & 0.7500 & 0.6667 & 0.6271 & 0.4627\\
  InternVL3.5-241B-A28B & \cmark & Open & 0.8944 & 0.9127 & 0.9291 & 0.9167 & 0.9153 & 0.8134\\
  Intern-S1 & \cmark & Open & 0.9014 & 0.9127 & 0.9223 & 0.9028 & 0.8814 & 0.7463\\
  Qwen3-VL-30B-A3B-Think & \cmark & Open & 0.8732 & 0.8810 & 0.9054 & 0.8843 & 0.9322 & 0.6940\\
  Qwen3-VL-Plus & \xmark & - & 0.8275 & 0.8968 & 0.8986 & 0.8565 & 0.9153 & 0.7687\\
  Qwen3-VL-8B-Think & \cmark & Open & 0.8768 & 0.8730 & 0.8885 & 0.9028 & 0.8983 & 0.6567\\
  Seed1.5-VL & \xmark & - & 0.9327 & 0.9127 & 0.9122 & 0.8472 & 0.8305 & 0.7015\\
  Qwen3-VL-235B-A22B-Instruct & \xmark & Open & 0.8204 & 0.8929 & 0.8986 & 0.8426 & 0.8814 & 0.7761\\
  InternVL3-78b & \xmark & Open & 0.8556 & 0.8730 & 0.8885 & 0.8981 & 0.9153 & 0.6194\\
  Qwen3-VL-4B-Think & \cmark & Open & 0.8838 & 0.8770 & 0.8615 & 0.9074 & 0.8983 & 0.6045\\
  Intern-S1-mini & \cmark & Open & 0.8239 & 0.8690 & 0.8547 & 0.8611 & 0.8475 & 0.6791\\
  GLM-4.1V-9B-Thinking & \cmark & Open & 0.8433 & 0.8690 & 0.8649 & 0.8657 & 0.8814 & 0.6493\\
  Qwen3-VL-32B-Instruct & \xmark & Open & 0.8169 & 0.8571 & 0.8885 & 0.8519 & 0.8305 & 0.6866\\
  Qwen2.5-VL-72B & \xmark & Open & 0.8063 & 0.8063 & 0.8770 & 0.9088 & 0.8102 & 0.9322\\
  Qwen2.5-VL-Max & \xmark & - & 0.7958 & 0.8571 & 0.8885 & 0.8194 & 0.8983 & 0.6642\\
  GPT-5-nano & \cmark & - & 0.8063 & 0.8452 & 0.8311 & 0.8241 & 0.7797 & 0.5672\\
  Qwen2.5-VL-32B & \xmark & Open & 0.7729 & 0.8413 & 0.8750 & 0.8009 & 0.8305 & 0.6418\\
  Gemini-2.5-Flash & \cmark & - & 0.7799 & 0.6111 & 0.6757 & 0.6620 & 0.7627 & 0.5373\\
  Qwen3-VL-8B-Instruct & \xmark & Open & 0.7113 & 0.8175 & 0.8446 & 0.8241 & 0.7627 & 0.5075\\
  Qwen3-VL-30B-A3B-Instruct & \xmark & Open & 0.7042 & 0.7937 & 0.8311 & 0.7824 & 0.7119 & 0.5970\\
  GPT-4o & \xmark & - & 0.7359 & 0.8175 & 0.7973 & 0.7500 & 0.7627 & 0.5224\\
  Qwen2.5-VL-7B & \xmark & Open & 0.6678 & 0.7659 & 0.8041 & 0.7130 & 0.6441 & 0.5373\\
  Qwen3-VL-4B-Instruct & \xmark & Open & 0.6708 & 0.7302 & 0.7804 & 0.7222 & 0.6610 & 0.5970\\
  Qwen3-VL-2B-Think & \cmark & Open & 0.7342 & 0.6706 & 0.7128 & 0.7083 & 0.6102 & 0.3657\\
  Qwen2.5-VL-3B & \xmark & Open & 0.6426 & 0.7381 & 0.7635 & 0.6898 & 0.6610 & 0.4776\\
  GPT-4o mini & \xmark & - & 0.6391 & 0.7302 & 0.7500 & 0.6667 & 0.6271 & 0.4627\\
  Qwen3-VL-2B-Instruct & \xmark & Open & 0.5405 & 0.6190 & 0.6318 & 0.6250 & 0.6102 & 0.3731\\
  Deepseek-VL2 & \xmark & Open & 0.4120 & 0.5040 & 0.4899 & 0.4907 & 0.3729 & 0.3060\\
  Random & - & - & 0.2500 & 0.2500 & 0.2500 & 0.2500 & 0.2500 & 0.2500\\
  \hline
  \end{tabularx}
  \begin{tablenotes}
      \footnotesize
      \item[a] Think: indicates models with inference-time reasoning capabilities
      \item[b] Weight: distinguishes between Proprietary (Closed) and Open-weight models
      \item[c] F.E.: Fact Extraction
      \item[d] RR\&F.: Reagent Roles \& Functions
      \item[e] M\&P.: Mechanism \& Process
      \item[f] CA\&R.: Comparative Analysis \& Reasoning
      \item[g] G.U.: Global Understanding
      \item[h] S.R.: Structure Recognition
  \end{tablenotes}
\end{threeparttable}
\end{table*}

\begin{table*}
\begin{threeparttable}
  \caption{Performance of MLLMs across FD-QA Task Types}
  \begin{tabularx}{\textwidth}{@{} l X X X X X X X @{}}
  \hline
  Model & Think\tnote{a} & Weight\tnote{b} & Rxn.En.\tnote{c} & Rxn.Zh.\tnote{d} & C.R.\tnote{e} & S.R.\tnote{f} & O.S.\tnote{g}\\
  \hline
  Gemini-3-Flash-preview & \cmark & - & \textbf{0.4574} & \textbf{0.4685} & 0.5530 & \textbf{0.4264} & \textbf{0.4630}\\
  Gemini-3-Pro-preview & \cmark & - & 0.4167 & 0.4407 & 0.5000 & 0.3897 & 0.4287\\
  Gemini-2.5-Pro & \cmark & - & 0.4148 & 0.4426 & \textbf{0.5682} & 0.3652 & 0.4287\\
  Gemini-2.5-Flash & \cmark & - & 0.3519 & 0.3815 & 0.4394 & 0.3235 & 0.3667\\
  Seed1.8-Think & \cmark & - & 0.3352 & 0.3889 & 0.4242 & 0.3064 & 0.3620\\
  GPT-5(high) & \cmark & - & 0.3444 & 0.3463 & 0.4470 & 0.3113 & 0.3454\\
  GPT-5.1(high) & \cmark & - & 0.3333 & 0.3426 & 0.4545 & 0.2941 & 0.3380\\
  Seed1.5-VL-Think & \cmark & - & 0.3148 & 0.3574 & 0.3939 & 0.2892 & 0.3361\\
  Qwen3-VL-32B-Think & \cmark & Open & 0.2981 & 0.3167 & 0.4015 & 0.2647 & 0.3074\\
  Qwen3-VL-235B-A22B-Think & \cmark & Open & 0.3056 & 0.3056 & 0.4091 & 0.2721 & 0.3056\\
  Qwen3-VL-30B-A3B-Think & \cmark & Open & 0.2611 & 0.2481 & 0.3409 & 0.2353 & 0.2546\\
  Qwen3-VL-8B-Think & \cmark & Open & 0.2333 & 0.2315 & 0.3333 & 0.2010 & 0.2324\\
  Qwen3-VL-4B-Think & \cmark & Open & 0.2352 & 0.1870 & 0.3106 & 0.2108 & 0.2111\\
  Intern-S1 & \cmark & Open & 0.1963 & 0.1944 & 0.2652 & 0.1740 & 0.1954\\
  Seed1.5-VL-Instruct & x & - & 0.1630 & 0.1741 & 0.2045 & 0.1495 & 0.1685\\
  Qwen3-VL-2B-Think & \cmark & Open & 0.1444 & 0.1463 & 0.1818 & 0.1324 & 0.1453\\
  GPT-4o & \xmark & - & 0.1426 & 0.1389 & 0.1515 & 0.1397 & 0.1407\\
  Qwen3-VL-235B-A22B-Instruct & \xmark & Open & 0.1241 & 0.1481 & 0.1742 & 0.1078 & 0.1361\\
  Qwen3-VL-32B-Instruct & \xmark & Open & 0.1241 & 0.0981 & 0.1894 & 0.1029 & 0.1111\\
  Qwen3-VL-8B-Instruct & \xmark & Open & 0.0907 & 0.1019 & 0.1591 & 0.0686 & 0.0963\\
  Qwen3-VL-30B-A3B-Instruct & \xmark & Open & 0.0778 & 0.0815 & 0.1288 & 0.0613 & 0.0796\\
  Qwen3-VL-4B-Instruct & \xmark & Open & 0.0611 & 0.0611 & 0.0606 & 0.0612 & 0.0611\\
  Qwen3-VL-2B-Instruct & \xmark & Open & 0.0481 & 0.0667 & 0.0758 & 0.0392 & 0.0574\\
  \hline
  \end{tabularx}
  \begin{tablenotes}
      \footnotesize
      \item[a] Think: indicates models with inference-time reasoning capabilities
      \item[b] Weight: distinguishes between Proprietary (Closed) and Open-weight models
      \item[c] Rxn.En.: RxnBench-Doc-En
      \item[d] Rxn.Zh.: RxnBench-Doc-Zh
      \item[e] C.R.: Context Reasoning (En)
      \item[f] S.R.: Structure Reasoning (En)
      \item[g] O.S.: Overall Score
  \end{tablenotes}
\end{threeparttable}
\end{table*}

\renewcommand{\lstlistingname}{}
\renewcommand{\thelstlisting}{S4}

\lstset{
  basicstyle=\ttfamily\small,
  breaklines=true,
  frame=single,
  columns=fullflexible,
  keepspaces=true,
  showstringspaces=false,
  escapeinside={(*@}{@*)},
  moredelim=**[is][\bfseries]{@b}{@}, 
}

\begin{figure*}[h]
\centering
\begin{lstlisting}[caption={Prompt used in SF-QA generation}]
@bRole@ :

Chemistry Expert (Organic \& Cheminformatics).

@bTask@ :

Analyze a reaction scheme image and generate one high-difficulty, reasoning-based MCQ in JSON format.

@bRequirements@ :

Content: Question must be grounded strictly in the image (SMILES for concrete, Extended-SMILES for Markush). Focus on mechanism, multi-step logic, or structural inference.

Options: 1 correct answer, 3 plausible but incorrect distractors. All must match in tone, length, and style.

Language: Provide both Chinese and English for all text fields.

Taxonomy: Select exactly one: 0 (Fact Extraction), 1 (Reagent Roles and Functions Identification), 2 (Reaction Mechanism and Process Understanding), 3 (Comparative Analysis and Reasoning), 4 (Multi-step Synthesis and Global Understanding), 5 (Chemical Structure Recognition).

JSON Output Only:

``` json

\{

"type": "X", \# from 0 to 5

"question\_zh": "question in Chinese",

"answer\_zh": "correct answer in Chinese",

"wrong\_options\_zh": {[}"wrong option 1 in Chinese", "wrong option 2 in
Chinese", "wrong option 3 in Chinese"{]},

"question\_en": "question in English",

"answer\_en": "correct answer in English",

"wrong\_options\_en": {[}"wrong option 1 in English", "wrong option 2 in
English", "wrong option 3 in English"{]}

\}

```

@bConstraints@ :

- No external knowledge.

- Single unambiguous answer.

- No markdown/commentary outside JSON.

- Do not use phrases like "as shown in the image".

\end{lstlisting}
\end{figure*}




























\end{document}